# Efficient Deployment and Mission Timing of Autonomous Underwater Vehicles in Large-Scale Operations

Somaiyeh MahmoudZadeh
School of Information Technology, Deakin University, Geelong, Victoria 3220, Australia
S.MahmoudZadeh@deakin.edu.au

**Abstract**

This study introduces a connective model of routing- local path planning for Autonomous Underwater Vehicle (AUV) time efficient maneuver in long-range operations. Assuming the vehicle operating in a turbulent underwater environment, the local path planner produces the water-current resilient shortest paths along the existent nodes in the global route. A re-routing procedure is defined to re-organize the order of nodes in a route and compensate any lost time during the mission. The Firefly Optimization Algorithm (FOA) is conducted by both of the planners to validate the model's performance in mission timing and its robustness against water current variations. Considering the limitation over the battery lifetime, the model offers an accurate mission timing and real-time performance. The routing system and the local path planner operate cooperatively, and this is another reason for model's real-time performance. The simulation results confirms the model's capability in fulfilment of the expected criterion and proves its significant robustness against underwater uncertainties and variations of the mission conditions.

## 1 Introduction

Autonomous Underwater Vehicles (AUVs) are designed to provide cost-effective underwater missions and largely used for different purposes over the past decades (Djapic and Nad, 2010). The problem associated with most of the todays AUV's autonomous operation is that they operate with a pre-defined mission outline and require human supervision, in which a set of pre-programmed instructions is fed to vehicle for any specific mission. Considering this deficiency, obtaining a premier autonomy to manage the mission time and autonomous adaption to the environmental changes is a substantial prerequisite in this regard. A vast literature exists on AUVs' routing and motion planning framework. Different deterministic algorithms, such as D* (Carsten et al., 2006), A* (Koay and Chitre, 2013), and FM* (Petres et al., 2007), have been used recently to address AUVs' motion planning problem. Deterministic approaches also have been investigated on vehicle's task allocation and routing problems, in which a multiple-target-multiple-agent framework based on graph matching algorithm has been studied by (Kwok et al., 2002). Both vehicle routing and path planning are categorized as a non-deterministic polynomial-time problem in which computational burden increases with enlargement of the problem search space. Hence, deterministic and heuristic algorithms cannot be appropriate for real-time applications as these methods are computationally expensive in large spaces (Roberge et al., 2013). Meta-heuristics are another alternative group of algorithms for solving complex problems that offer near optimal solutions in a very quick computation (Atyabi et al., 2018; MahmoudZadeh et al., 2017-*a*) and is appropriate for the purpose of this study.

There are various examples of evolution-based applications of path planning and routing-scheduling approaches. A Non-Dominated Sorting Genetic Algorithm (NSGA-II) is employed for AUV's waypoint guidance and offline path planning (Ataei and Yousefi-Koma, 2014). MahmoudZadeh et al., designed an online Differential Evolution (DE) based path planner for a single AUV's operation in a dynamic ocean environment (MahmoudZadeh et al., 2018-*a*). A routing-task-assigning framework is also introduced recently for an AUV's mission planning in a large static operating network, in which the performance of genetic algorithm, imperialist competitive algorithm, and Particle Swarm Optimization (PSO) methods are tested and compared in solving the graph complexity of the routing problem (MahmoudZadeh et al., 2015). Afterward, they extended their study by modelling a more complex environment where a semi-dynamic operation network is encountered in contrast and subsequently efficiency of the biogeography-based optimization and PSO algorithms are tested and evaluated in solving the dynamic routing and task allocation approach (MahmoudZadeh et al., 2016-*a,c*).

Indeed, attaining a superior optimization and computationally efficient approach for addressing these complex problems is still an open area for further investigation. Assuming a waypoint cluttered graph-like environment, the AUV must be able to manage its battery lifetime to carry out a mission including specific set of waypoints; hence, a general route planning over the operation network is primary requirement for this purpose. The second essential objective is to adapt the ocean current deformations and safely guide the AUV trough the network vertices. To do so, the system should be computationally efficient to take a real-time trend over the subsea current deformations. Current research constructs a general routing system with a mounted local path planner to provide a reliable and energy efficient maneuver for the AUV. This system takes the meta-heuristics advantages of Firefly Optimization Algorithm (FOA) to meet the requirements of a long-range operation in a turbulent subsea environment. This research conducts a two dimensional turbulent current map generated by a popular predictive model based on superposition of multiple Lamb vortices (Garau et al., 2006; MahmoudZadeh et al., 2017-*b* ; 2018-*c*-*e*).

# 2 Routing Problem in a Waypoint Cluttered Environment

The operation space is modelled as an undirected weighted graph ($G$) including a specific number of nodes denoted by $P$ and graph connections/edges ($E$). The vertices of the network $p^{i}_{xyz}\in P$ are uniformly distributed in a three dimensional volume of ($x_{10000}$,$y_{10000}$,$z_{100}$) that represented as follows:

$$G(P,E)\xrightarrow[|E|=m]{|P|=k}\begin{cases}P:\left\langle p^1,\dots,p^k\right\rangle\\E:\left\langle e^1,\dots,e^m\right\rangle\\e^{ij}=\left(p^i,p^j\right)\end{cases}\tag{1}$$

Any edge between $p^i$ and $p^j$ in the graph ($e^{ij}$) has a corresponding length of ($l_{ij}$) and approximated traversing time, given by (2). In the given operating graph, the AUV should meet maximum possible nodes in a restricted battery lifetime. Accordingly, the route planner tends to determine a best set of nodes in the graph to guide the AUV toward the target node and to accommodate battery restriction. With respect to given definitions, a route ($\Re$) is mathematically indicated as follows:

$$\begin{aligned}&\forall e^{ij}\quad\exists\ l_{ij},t_{ij}\\&l_{ij}=\left((p_x^j-p_x^i)^2+(p_y^j-p_y^i)^2+(p_z^j-p_z^i)^2\right)^{\frac{1}{2}}\\&t_{ij}={l_{ij}}\Big/{|\upsilon|}+\delta_{ij}\end{aligned}\tag{2}$$

$$\begin{aligned}&e^{ij}:\left\langle p_{xyz}^i,p_{xyz}^j\right\rangle\Rightarrow\Re=\sum_{\substack{i=1\\j\neq i}}^{|E|}S\times e^{ij};\quad S=\{0,1\}\\&\Re:\left\langle\underbrace{e^{si},\dots,e^{ik},\dots,e^{kj},\dots e^{jt}}_{\ell\subseteq\{1,\dots,|E|\}}\right\rangle;\quad\begin{matrix}e^{si}:\left\langle p_{xyz}^s,p_{xyz}^i\right\rangle\\e^{jt}:\left\langle p_{xyz}^j,p_{xyz}^t\right\rangle\end{matrix}\\&\mathrm{T}_\Re=\sum_{\substack{i=1\\j\neq i}}^{|E|}S\times e^{ij}\times t_{ij}=\sum_{\substack{i=1\\j\neq i}}^{|E|}S\times e^{ij}\left(l_{ij}\times|\upsilon|^{-1}\right)\end{aligned}\tag{3}$$

here, $\ell$ is the length of the route which is subset of total number existent edges in the graph ($|E|$). $\upsilon$ denotes the vehicle's water referenced velocity in the body frame. $S$ is a selection variable that represents selection of any arbitrary edge in the graph. $\mathrm{T}_\Re$ is the route time from start node of $p^s_{xyz}$ to target node of $p^t_{xyz}$. The battery lifetime denoted by $T_\tau$ and is started to counting inversely from the beginning of the operation. The $\mathrm{T}_\Re$ should approach the $T_\tau$ but should not overstep that. The route should not include non-existent edges, and should not traverse a specific edge for multiple times.

# 3 Environmental Dynamics and Local Path Planning

In order to deal with environmental impact on vehicles motion, a local path planner is conducted in this study to operate in a smaller operating window between pairs of route nodes. This space reduction leads reducing the computation burden as a smaller window is required to be monitored. Water current is an important environmental factor that influences AUV's motion. The local path planner aims to find a time efficient path while accommodating the current deformations. The current map data in this research is obtained from a popular numerical estimation model based of recursive Navier-Stokes equations (Garau et al., 2006) as follows:

$$\upsilon_c=\left(\upsilon_{c,x},\upsilon_{c,y}\right)\Rightarrow\begin{cases}\upsilon_{c,x}=|\upsilon_c|\cos\theta_c\cos\psi_c\\\upsilon_{c,y}=|\upsilon_c|\cos\theta_c\sin\psi_c\end{cases}\tag{4}$$

here, the $\upsilon_c$ is current velocity vector and the $\upsilon_{c,x}$ and $\upsilon_{c,y}$ are the $x$-$y$ components of the $\upsilon_c$. The physical model used by the AUV to diagnose the current velocity field can be found in (Garau et al., 2006;

MahmoudZadeh et al., 2018-*b*-*d*). AUV's motion in six degree of freedom is provided by state variables of body and NED frames (Fossen 2002), as follows:

$$\begin{aligned}&\eta:(X,Y,Z,\varphi,\theta,\psi)\\&\upsilon:(\upsilon_x,\upsilon_y,\upsilon_z,p,q,r)\end{aligned} \tag{5}$$

where, the $\eta$ and $\upsilon$ denote vehicle's dynamics and kinematic over the time. $X,Y,Z$ denote AUV's position along the path. $\varphi,\theta,\psi$ are the Euler angles of roll, pitch, and yaw, respectively. The $\upsilon$ is AUV's velocity vector in the body frame; $\upsilon_x$,$\upsilon_y$,$\upsilon_z$ are directional velocities of surge, sway and heave; and $p$,$q$,$r$ are the rotational velocities. In this study, the local path $\wp$ is generated using B-Spline curves captured from number of control points while the water current velocity is continuously taken into account. The local path curve $\wp$ is calculated by:

$$\begin{aligned}&\theta_t=\tan^{-1}\left(-\left|\Delta Z_{i,t}\right|\Big/\sqrt{\Delta X_{i,t}{}^2+\Delta Y_{i,t}{}^2}\right)\\&\psi_t=\tan^{-1}\left(\left|\Delta Y_{i,t}\right|\big/\left|\Delta X_{i,t}\right|\right)\\&\upsilon_{x,t}=|\upsilon|\cos\theta_t\cos\psi_t+|\upsilon_c|\cos\theta_c\cos\psi_c\\&\upsilon_{y,t}=|\upsilon|\cos\theta_t\sin\psi_t+|\upsilon_c|\cos\theta_c\sin\psi_c\\&\upsilon_{z,t}=|\upsilon|\sin\theta_t\\&\wp=[X,Y,Z,\psi,\theta,\upsilon_x,\upsilon_y,\upsilon_z]\end{aligned} \tag{6}$$

The AUV is presumed with a constant thrust power; hence, the path time $T_\wp$ has a linear relation to path length. The water current deviates the vehicle from its desired trajectory; hence, the resultant path should meet the kinematic constraints of the vehicle in dealing with current force. Therefore, AUV's surge-sway velocities and its yaw-pitch orientation should be constrained to $\upsilon_{x,\max}$, [$\upsilon_{y,\min}$, $\upsilon_{y,\max}$], $\theta_{\max}$, and [$\psi_{\min}$, $\psi_{\max}$] in all states along the path. Accordingly, the path cost is calculated by (7).

$$\begin{aligned}&\forall\,\wp^i_{x,y,z}\\&T_\wp=\sum_{i=p^a_{x,y,z}}^{|\wp|}\left(\Delta X_{i,t}{}^2+\Delta Y_{i,t}{}^2+\Delta Z_{i,t}{}^2\right)^{1/2}\times\left(|\upsilon|\right)^{-1}\\&C_\wp=T_\wp+\varepsilon_{\upsilon_x}\max\left(0;\upsilon_{x,t}-\upsilon_{x,\max}\right)+...\\&\quad...+\varepsilon_{\upsilon_y}\max\left(0;\left|\upsilon_{y,t}\right|-\upsilon_{y,\max}\right)+...\\&\quad...+\varepsilon_\theta\max\left(0;\theta_t-\theta_{\max}\right)+...\\&\quad...+\varepsilon_\psi\max\left(0;\left|\dot{\psi}_t\right|-\psi_{\max}\right)\end{aligned} \tag{7}$$

The $\varepsilon_{\upsilon x}$, $\varepsilon_{\upsilon y}$, $\varepsilon_\theta$, $\varepsilon_\psi$ denote the impact of each constraint violation in determination of the local path cost $C_\wp$.

# 4 Mission Evaluation Criterion

The generated route ($\Re$) is composed of distances between nodes ($l_{ij}$) and the path planner generates time efficient trajectory along those distances ($l_{ij}\propto\wp_{ij}$); hence, the path cost of $C_\wp$ directly impacts the route cost of $C_\Re$. As mentioned earlier in Section II, the rout time $T_\Re$ should approach the total battery lifetime $T_\tau$, but should not overstep that. Therefore, the $C_\Re$ gets penalty when the $T_\Re$ for a particular route exceeds the $T_\tau$. The local path may take longer time in dealing with environmental dynamic changes. In such a case, the lost time should be compensated by a proper re-routing process. Consequently, re-routing computation cost is considered in total mission cost calculation. Thus, the $C_{\mathfrak{R}}$ and total mission cost of $C_{\tau}$ in the proceeding research is calculated by (8).

$$
\begin{aligned}
&C_{\wp_{ij}} \approx \mathrm{T}_{\wp}^{ij} \propto t_{ij} \Rightarrow C_{\wp_{ij}} \propto t_{ij} \\
&\mathrm{T}_{\mathfrak{R}} = \sum_{\substack{i=0 \\ j \neq i}}^{|E|} S.e^{ij} \times t_{ij} \\
&C_{\mathfrak{R}} = \left| \mathrm{T}_{\mathfrak{R}} - \mathrm{T}_{\tau} \right| \times \max\left( 0, {}^{\mathrm{T}_{\mathfrak{R}}}\!/\!_{\mathrm{T}_{\tau}} \right) \\
&C_{\mathfrak{R}} = \left| \sum_{\substack{i=0 \\ j \neq i}}^{|E|} S.e^{ij} \times \left( C_{\wp_{ij}} + \delta_{\wp_{ij}} \right) - \mathrm{T}_{\tau} \right| \times \max\left( 0, \mathrm{T}_{\mathfrak{R}} \mathrm{T}_{\tau}^{-1} \right) \\
&C_{\mathfrak{R}} \propto f\left( C_{\wp_{ij}}, \mathrm{T}_{\tau} \right) \\
&C_{\tau} = C_{\mathfrak{R}}\left( C_{\wp_{ij}}, \mathrm{T}_{\tau} \right) + \sum_{1}^{r} T_{compute}
\end{aligned}
\qquad (8)
$$

Where, $T_{compute}$ is the re-routing computation time, and $r$ is the number of re-routing in a mission. $\delta_{\wp ij}$ is the delayed time during the local path planning between $p^{i}_{xyz}$ and $p^{j}_{xyz}$.

# 5 FOA on Mission Routing and Path Planning

Firefly Optimization Algorithm is a meta-heuristic algorithm inspired from the flashing patterns of fireflies, in which the fireflies attract each other based on their brightness (Yang 2010). The fireflies' brightness decreases by distance and the brighter fireflies attract the less bright ones; hence, their attraction is proportional to their brightness and their relative distance. Attraction of a firefly $i$ toward the brighter firefly $j$ is calculated as follows:

$$
\begin{aligned}
&\partial_{ij} = \left\| \chi_j - \chi_i \right\| \\
&\chi_{i,t+1} = \chi_{i,t} + \beta_0 e^{-\gamma \partial_{ij}^{2}} (\chi_{j,t} - \chi_{i,t}) + \alpha_t \varsigma_i^t \\
&\alpha_t = \alpha_0 \kappa^t, \ \ \kappa \in (0,1)
\end{aligned}
\qquad (9)
$$

the $\partial_{ij}$ is the distance between fireflies $i$ and $j$; $\beta_0$ is the attraction factor at $\partial$=0, $\alpha_0$ and $\alpha_t$ are the initial randomness scaling value and the randomization parameter, respectively. $\alpha_t$ tunes the randomness of fireflies' movement in each iteration. $\kappa$ is a damping factor. The $\varsigma_i^t$ is a randomly generated vector at time $t$. The $\gamma$ light absorption factor. In a case that $\beta_0$ approaches zero the movement turns to a simple random walk, while $\gamma$= 0 turns the FOA to a variant of PSO; thus, a proper balance should be set between the engaged parameters (Yang 2010). The FOA is efficient due to applying an automatic subdivision approach that enhances convergence rate of the algorithm, and iteratively prevents fireflies from trapping into local optima. This accommodates FOA to efficiently deal with highly nonlinear continuous problems, and makes it flexible in dealing with multimodality (Yang and He, 2013). The control parameters in FOA can be tuned iteratively, which is another reason for its fast convergence. Similar to other metaheuristic algorithms, the FA also has two inner loops through the population $i_{\max}$ and iteration $t_{\max}$, so at the extreme case the algorithms complexity is O($i_{\max}^{2} \times t_{\max}$); hence, the computation cost is respectively low as its complexity is linear to time. The cost evaluation is the most computationally complex part of almost all optimization problems. To the purpose of AUV global routing, first step in using the FOA algorithm is to provide the initial population in the format of feasible routes, which has a great impact on algorithms performance. Fireflies in this context are defined as feasible routes in the graph (MahmoudZadeh et al., 2016-*b,* 2019).

**FOA based Route-Path Planning/Re-planning**

Initialization phase for the **Route Planning System**:

- Set the maximum number of iteration $t_{max}$ and the population size $i_{max}$.
- Construct global routes $\Re_i$ according to adjacency connections and randomly generated priority vector.
- Initialize solution (Firefly) vectors $\chi^{\Re}_{i,t}$ with the generated route vectors $\Re_i$
- Define light absorption and attraction coefficients for routes $\gamma_{\Re}$, $\beta_{\Re 0}$
- Define the randomness scaling factor of $\alpha_{\Re 0}$
- Set the damping and randomization factors of $\kappa_{\Re}$ and $\alpha_{\Re}$

***For*** $t$ =1 ***to*** $t_{max}$
  Evaluate each candidate Firefly $\chi^{\Re}_{i,t}$ according to route cost function $C_{\Re}$
  ***For*** $i$ =1 ***to*** $i_{max}$
    Reconstruct the route $\Re$ according to firefly $\chi^{\Re}_{i,t}$
    Evaluate the $\Re$ by $C_{\Re}(\chi^{\Re}_{i,t})$
    ***For*** $j$ =1 ***to*** $i$
      Reconstruct the route $\Re$ according to firefly $\chi^{\Re}_{j}$
      Evaluate the $\Re$ by $C_{\Re}(\chi^{\Re}_{j,t})$
      Update light intensity of $\chi^{\Re}_{i,t}$:
      $\partial_{ij} = \|\chi^{\Re}_{j,t} - \chi^{\Re}_{i,t}\|$
      $\beta_{\Re} = \beta_{\Re_0} e^{-\gamma_{\Re}\partial^2_{ij}}$
      $\chi^{\Re}_{i,t+1} = \chi^{\Re}_{i,t} + \beta_{\Re}(\chi^{\Re}_{j,t} - \chi^{\Re}_{i,t}) + \alpha_{\Re,t}\,\varsigma_{i,t}$
        ***if*** $(\beta\chi^{\Re}_{j,t} > \beta\chi^{\Re}_{i,t})$,
          Move firefly $\chi^{\Re}_{i,t}$ towards $\chi^{\Re}_{j,t}$
          ***else***
          Move firefly $\chi^{\Re}_{j,t}$ towards $\chi^{\Re}_{i,t}$
        ***end*** **(*if*)**
    ***end*** **(*For*)**
  ***end*** **(*For*)**
  Rank the fireflies and find the current best
***end*** **(*For*)**
Output the corresponding optimum route $\Re$
Output the corresponding route time $T_{\Re}$
Output the waypoint sequence
Output expected time ($t_{ij}$) for passing edge $e_{ij}$
Send these information to local path planner

**Shift Control to the Local Path Planner**

Initialization phase for the local path planner:

- Set the maximum iteration $t_{max}$ and population of $i_{max}$
- Initialize fireflies $\chi^{\wp}_{i,t}$ with the control points along the path $\wp_{xyz}$
- Set the parameters of: $\gamma_{\wp}$, $\beta_{\wp 0}$, $\alpha_{\wp 0}$, $\kappa_{\wp}$ and $\alpha_{\wp}$

***For*** $t$ =1 ***to*** $t_{max}$
  ***For*** $i$ =1 ***to*** $i_{max}$
    Reconstruct the route $\wp$ according to firefly $\chi^{\wp}_{i,t}$
    Evaluate the $\wp_i \approx \chi^{\wp}_{i,t}$ by $C_{\wp}(\chi^{\wp}_{i,t})$
    ***For*** $j$ =1 ***to*** $i$
      Reconstruct the route $\wp$ according to firefly $\chi^{\wp}_{j,t}$
      Evaluate the $\wp_j \approx \chi^{\wp}_{j,t}$ by $C_{\wp}(\chi^{\wp}_{j,t})$
      Update light intensity of $\chi^{\wp}_{i,t}$
      $\partial_{ij} = \|\chi^{\wp}_{j,t} - \chi^{\wp}_{i,t}\|$
      $\beta_{\wp} = \beta_{\wp_0} e^{-\gamma_{\wp}\partial^2_{ij}}$
      $\chi^{\wp}_{i,t+1} = \chi^{\wp}_{i,t} + \beta_{\wp}(\chi^{\wp}_{j,t} - \chi^{\wp}_{i,t}) + \alpha_{\wp,t}\,\varsigma_{i,t}$
      ***if*** $(\beta\chi^{\wp}_{j,t} > \beta\chi^{\wp}_{i,t})$,
        Move firefly $\chi^{\wp}_{i,t}$ towards $\chi^{\wp}_{j,t}$
        *else*
        Move firefly $\chi^{\wp}_{j,t}$ towards $\chi^{\wp}_{i,t}$
      ***end*** **(*if*)**
    ***end*** **(*For*)**
  ***end*** **(*For*)**
  Rank the fireflies and find the current best
***end*** **(*For*)**
Output result the best produced path $\wp$
Output the corresponding path time $T_{\wp}$

**Shift Control to the Re-Planner**

**INPUT:** $\wp$ and $T_{\wp}$ from the local path planner
**INPUT:** $\Re$, $T_{\Re}$, and $t_{ij}$ from the global route planner

***While*** (Current Waypoint ≠ Destination)
  ***if*** $(T_{\wp}^{ij} \le t_{ij})$,
    Continue the current optimum global route $\Re$
    Updated total available time $T_r = T_r - T_{\wp}^{ij}$
    Send the next pair Waypoints to the path planner.
    Shift the control to the path planner.
  ***else if*** $(T_{\wp}^{ij} > t_{ij})$,
    Replan Flag ==1
    Eliminate the visited edges from the graph
    Set the current waypoint as a new start point
    Send the new adjacency information to the global route planner
    Updated total available time $T_r = T_r - T_{\wp}^{ij}$
    Sent the updated $T_r$ to the global route planner
    Shift the control to the global route planner
  ***end*** **(*if*)**
  Calculate the total mission cost $C_r$
***end*(*While*)**

Figure 1. Pseudocode of FOA-based routing, path planning, and re-planning

The solutions take variable length limited to number of vertices in the graph that are generated using graph adjacency information. Accordingly, the algorithm stars to optimize the solutions based on defined cost function for routing problem. In the case of local path planning, the fireflies in the initial population are assigned with candidate local path solutions that are generated by a set of B-Spline control points. Then the FOA tends to efficiently locate the control points of a candidate $\wp$ curve in the solution space according to the defined cost function for the local path. The FOA process of AUV routing, path planning and re-planning is provided by a pseudo-code in Fig.1.

The battery lifetime $T_\tau$ should be managed adaptively. Accordingly, the local path time $T_\wp^{ij}$ gets compared to expected path time of $t_{ij}$ after visiting each node in the route sequence and if it exceeds that, re-routing flag gets triggered. The $T_\tau$ gets updated simultaneously. The given process in the pseudo code of Fig.1 continues until the AUV reaches to the target node.

# 6 Discussion on Simulation Results

First we turn to evaluate the performance of FOA-based local path planner according to given cost function in (7). The vehicle is assumed to move with a standard thrust power of maximum $\upsilon$=5.5 (*knots*). The battery consumption for a path is a constant multiple of the path time and path length due to proportional relation of current velocity to the cube root of the thrust. A static current map data is used to evaluate the behaviour of local path planner to water currents deformations. The current map is generated using a Gaussian distribution of 11 vortices in 100×100 grid. The paths' curvature is acquirable by the AUV's directional velocity components and radial acceleration. Figure.2 represents the local path behavior with respect to water current flow.

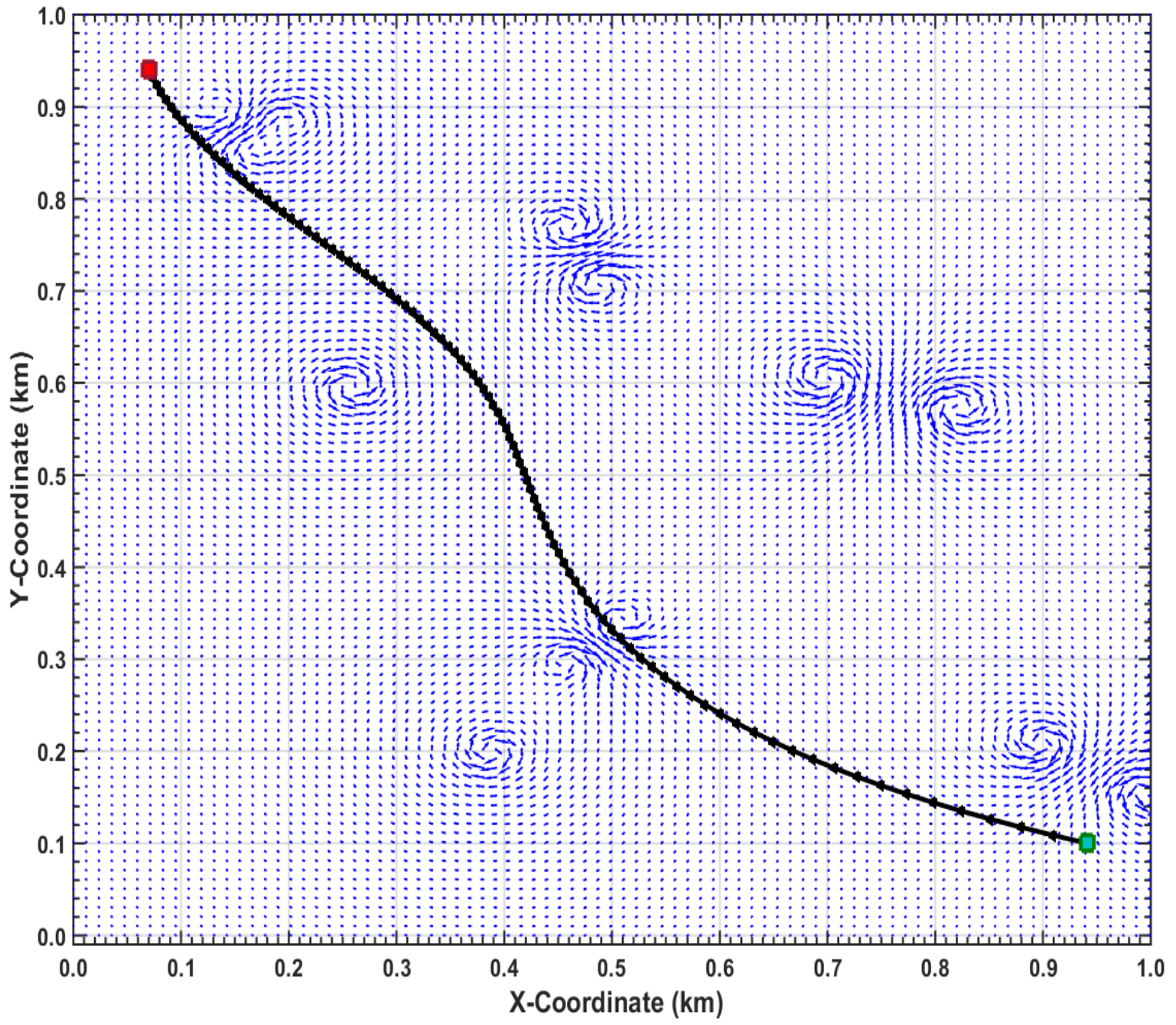

Figure 2. The local path adaption to current arrows in a static map

As depicted in Fig.2, it is noteworthy to hint the efficient capability of the FOA-based planner in conforming current arrows either in using accordant current arrows or in avoiding turbulent (vortices). According to path cost function, the path planner aims to determine the shortest battery efficient path between nodes and adapting water current deformations while the actuators boundary conditions and vehicular constraints are considered.

With respect to (7), the path cost function gets penalty when the generated path is violated the boundaries on vehicle's surge, sway, theta rate, yaw rate constraints, which here is defined as follows: $\upsilon_{x,\max}$=5.25 (*knots*); $[\upsilon_{y,\min},\upsilon_{y,\max}]$=[-0.97,0.97] (*knots*); $\theta_{max}$=20 (*deg/s*); and $[\psi_{\min},\psi_{\max}]$=[-17,17] (*deg/s*). Figure.3 presents the local path planner's performance in reducing the path cost and satisfying the abovementioned constraints.

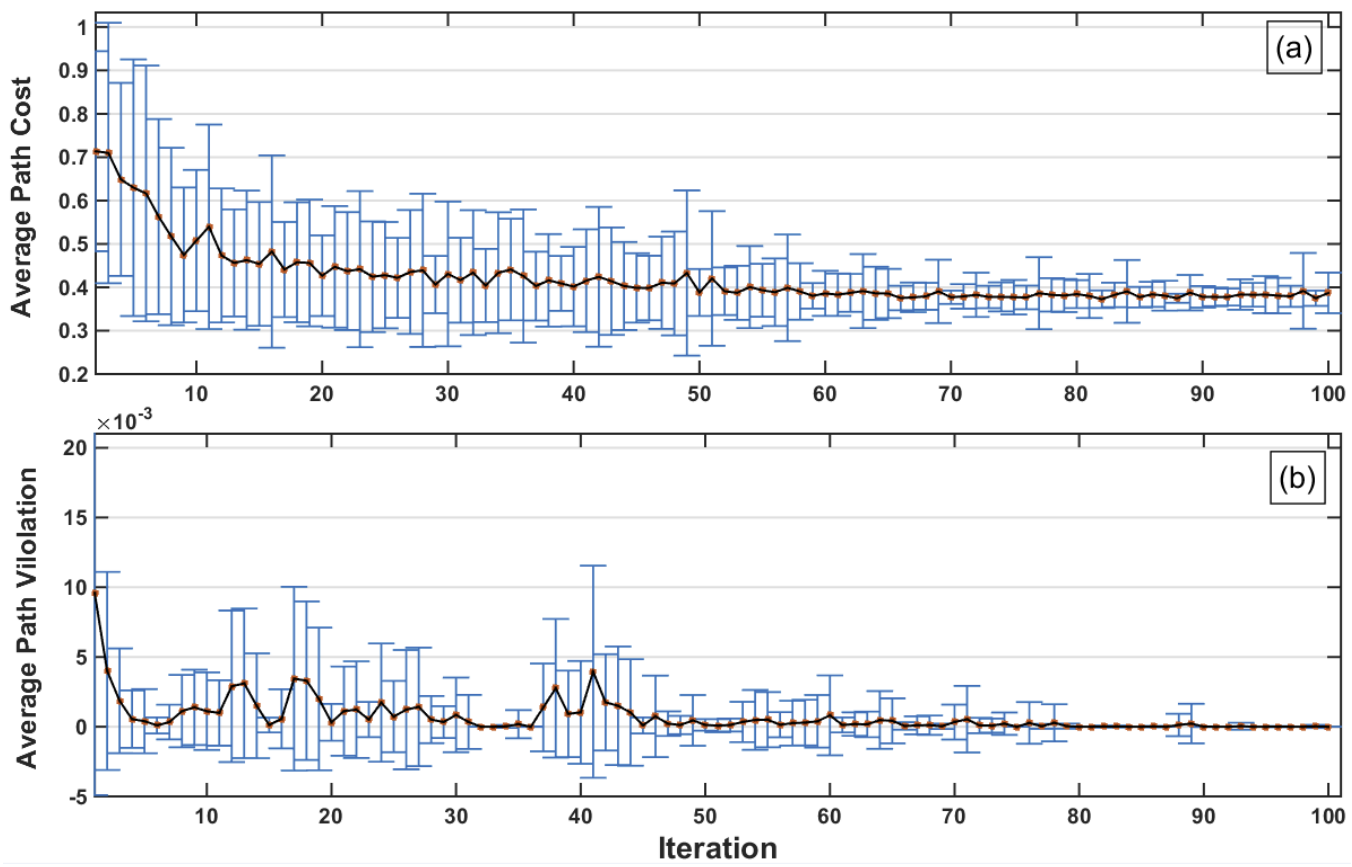

Figure 3. **(a)** Cost variations of path population over 100 iterations;
**(b)** Path violation of $\upsilon_x$, $\upsilon_y$, $\theta$, and $\psi$ over 100 iterations;

The generated path, as illustrated in plot Fig.3, shows a great fitness regarding all defined path constraints. The cost variation of path population experiences a moderate convergence to the minimum cost and the variation range narrows down iteratively. It is further outstanding from Fig.3 (b), the FOA-based path planner accurately manages the path toward eliminating the violation factors as the violation of the path population diminishes over the 100 iterations.

On the other hand, the routing model should select an efficient set of nodes restricted to battery life time $T_\tau$ to ensure on-time mission termination. A critical factor for concurrency of the routing and path planning models is having a short computational time to keeps any of them from dropping behind the process of the other one. Fig.4 presents the computational performance of the both FOA-based route planner and path planner in 25 simultaneous runs. Moreover, compatibility of the expected time $t_{ij}$ and the path time $T_{\wp}$ for traversing $l_{ij}$ is another significant performance metric impacts the system synchronism. Hence, there should not be a huge difference between variations of these two parameters. This concurrency also impacts on-time re-routing procedure. The concurrency of $t_{ij}$ and $T_{\wp}$ in 25 experiments is depicted by Fig.5. Routing and path planning computational time variations, as presented in Fig.4, are fairly drawn in a narrow range of seconds for all 25 experiments, which hint the real-time performance of the proposed connective FOA-based model in handling the environmental changes.

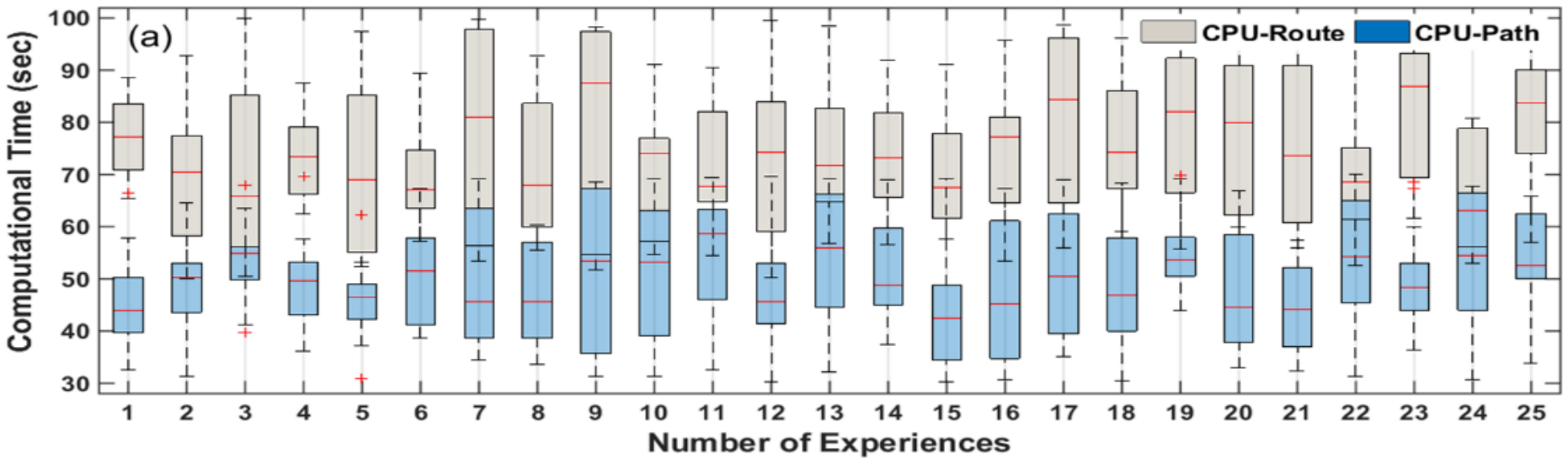

Figure 4. Computational time variation of route-path planning model over the 25 experiments

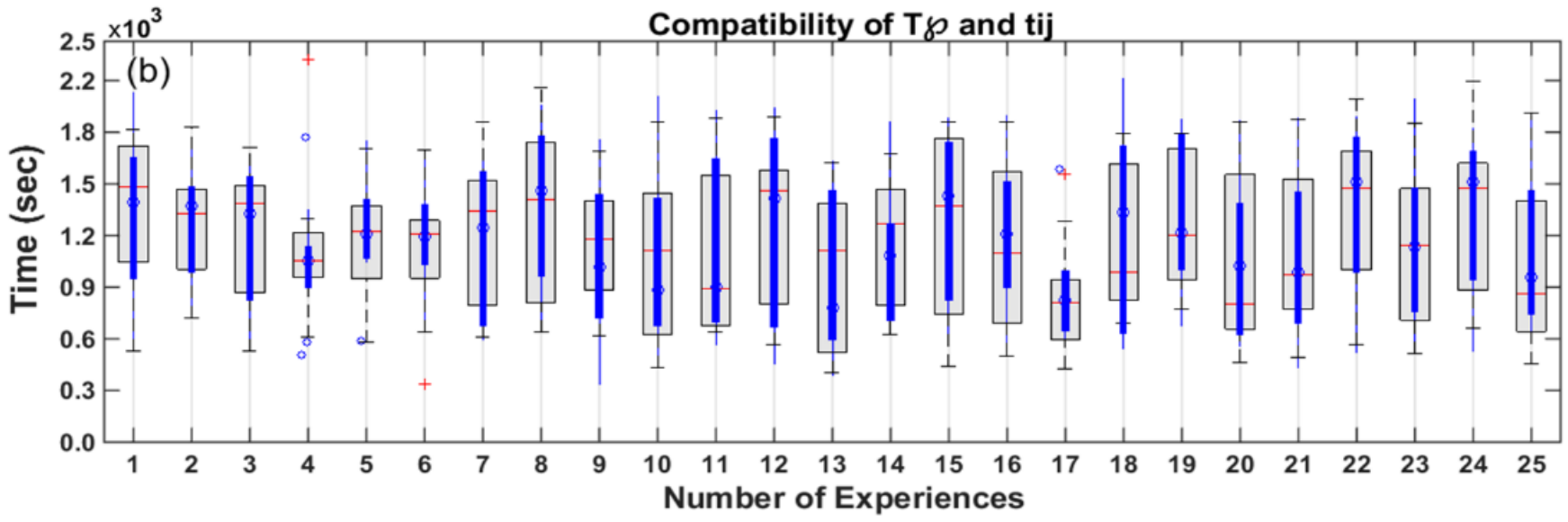

Figure 5. Compatibility of the value of $T_{\wp}$ and $t_{ij}$ in a quantitative manner over the 25 experiments

Analysis of the captured result from multiple experiences, indicates model's consistency in preserving the conformity between $t_{ij}$ (depicted by gray transparent box plot) and $T_{\wp}$ (depicted by blue compact box plot) as their average variations is relatively close in each experiment. This confirms the accurate synchronization of the routing and path planning system. The whole process of one experiment is illustrated by Fig.6 for better understanding, in which this single mission includes three re-routing and 11 local path planning passing through the 12 nodes. The routing system provides an initial efficient route. The remained time is initialized with battery life time $T_{\tau}$ and is counted inversely during the mission. The local path planner incorporates local environmental changes and if the $T_{\wp}$ oversteps the $t_{ij}$ the re-routing flag is triggered and controller shifts to the routing system to compensate the lost time.

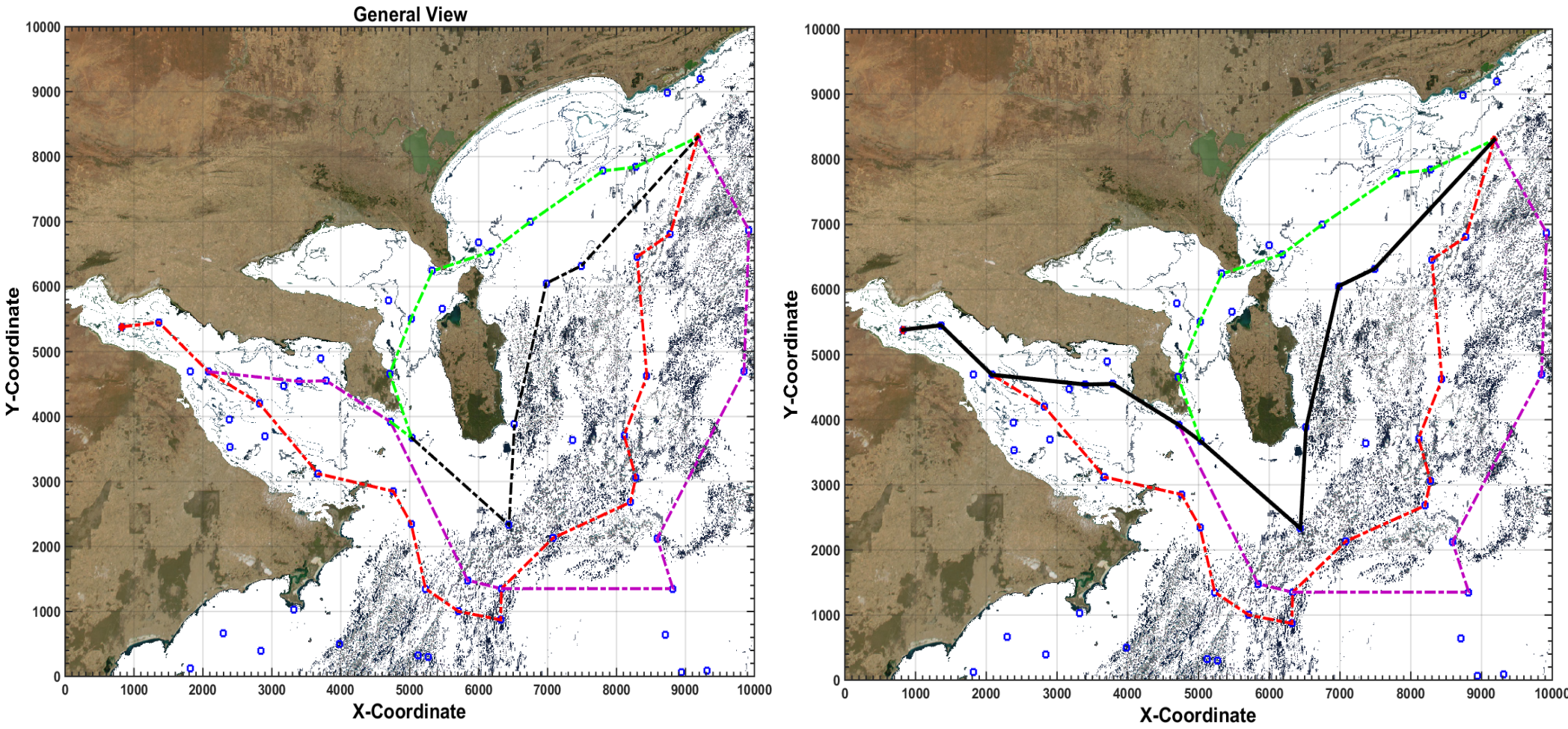

Figure 6. Routing, path planning, and re-routing procedure by re-arrangement of edges' order in a single mission.

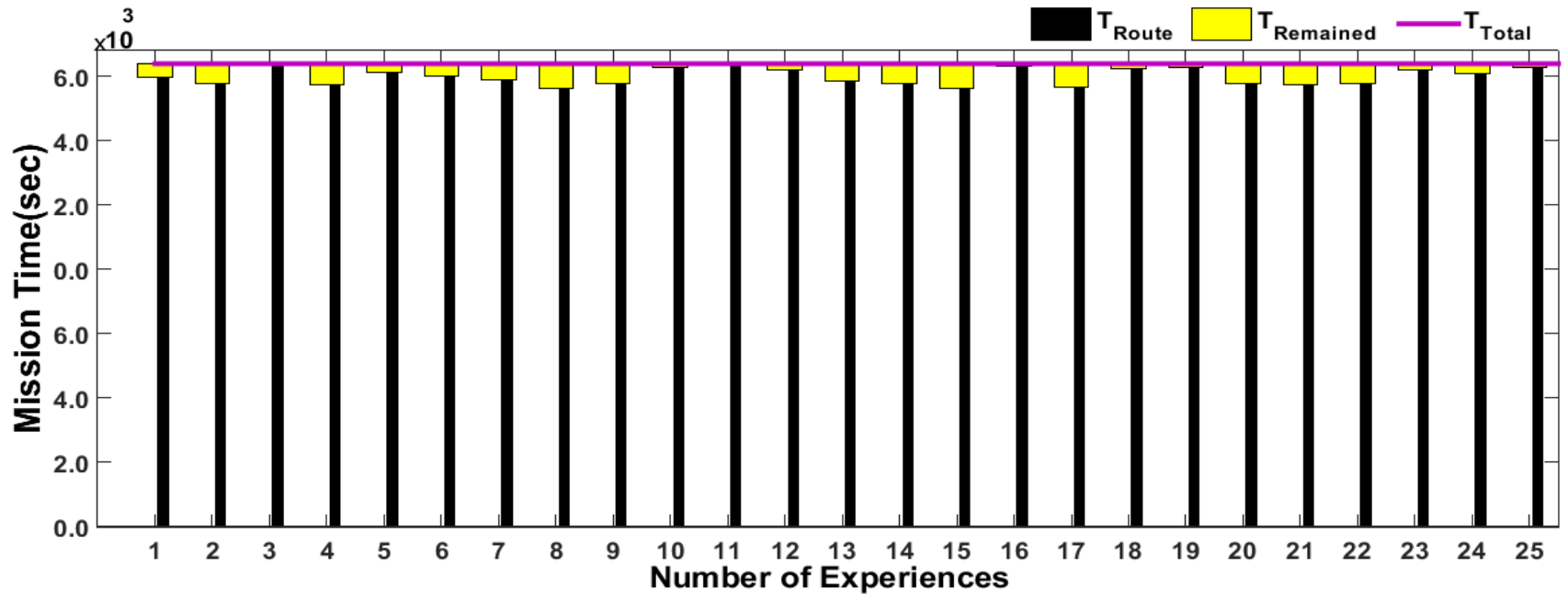

Figure 7. Statistical analysis of the model's timing performance in 25 missions.

As presented in Fig.6, the final optimum route (black line) is generated through the three re-planning process, in which the first route (presented by dashed red line) is discarded after passing two nodes; the second one (shown by pink dashed line) is discarded after visiting 5 nodes from the starting point, and the third route (depicted by green dashed line) is discarded in the node 6. These re-planning are carried out to compensate the lost time in the local path planning process.

The most important performance metric in this study is model's accuracy in mission timing and ensuring on-time completion of the mission. Thus, the best outcome of the model is to take a maximum use of battery life time and to fulfill a mission with minimum residual time. The model's capability of mission timing is examined through the 25 individual experiments (missions) presented by Fig.7, in which the battery life time is set on $T_\tau$=7.2 $\times 10^3$(*sec*) and the terrain is modelled as a realistic underwater environment encountering static ocean current map.

It is outstanding from Fig.7, the $T_{remained}$ is positive and it is approached to zero in all 25 missions, which means all missions completed before vehicle runs out of battery. Accordingly, the mission time ($T_{Route}$) maximized to approach upper bound of $T_{Total}$ (presented by pink horizontal line in Fig.7), but it doesn't overstep the line in any of experiments. It is noted from analyzing the results, the model accurately satisfies mission timing constraints along with other considerations. This is a significant achievement toward having a successful and reliable operation through the excellent mission time management.

# 7 Conclusion

In this study a connective model of AUV routing and local path planning based on firefly optimization algorithm (FOA) is presented, in which the model is advantaged with a reactive re-routing capability that manages the mission time by re-organizing the order of nodes in a way to be fitted to the battery life time. The local path planner, at the same time, tends to generate energy/time efficient paths along the selected nodes in a route encountering desirable and adverse water current flow. To validate the proposed connective model, the vehicle's operation is simulated in large-scale three-dimensional volume and the static water current map is added to consideration. The FOA performance on the proposed model is tested through the 25 individual mission trials. It is inferred from simulation results that the offered connective model proposes an efficient computational performance (in range of seconds) for both vehicle routing and local path planning that affirms the real-time performance of the model in long range mission management. The local planner also shows a great current resilient efficiency that leads remarkable energy saving in vehicle's continuous deployments. As inferable from the simulation results, the re-planning facilitates the vehicle to have a reliable and energy efficient operation by having an excellent mission timing. The future research will concentrate on expanding the proposed model in terms of upgrading the planners' capabilities and environmental influences on small and long-range missions. It is planned to expand the current study and to prepare a full version as a journal paper.

# References

Ataei, M., Yousefi-Koma, A. (2014) *Three-dimensional optimal path planning for waypoint guidance of an autonomous underwater vehicle*. Robotics and Autonomous Systems, 2014

Atyabi, A., MahmoudZadeh, S., Nefti-Meziani S. (2018) Current Advancements on Autonomous Mission Planning and Management Systems: an AUV and UAV perspective. ***Journal of Annual Reviews in Control*** - Elsevier, 46, 196-215

Carsten, J., Ferguson, D., Stentz, A. (2006) *3D field D*: improved path planning and replanning in three dimensions*. IEEE International Conference on Intelligent Robots and Systems, 2006, pp.3381-3386.

Djapic, V., Nad D. (2010) *Using collaborative autonomous vehicles in mine countermeasures.* Oceans'10 IEEE Sydney, 2010.

Fossen, T. (2002) *Marine Control Systems: Guidance, Navigation and Control of Ships, Rigs and Underwater Vehicles*. Marine Cybernetics Trondheim, Norway, 2002.

Garau, B., Alvarez, A., Oliver, G. (2006) *AUV navigation through turbulent ocean environments supported by onboard H-ADCP*. IEEE International Conference on Robotics and Automation, Orlando, Florida, May, 2006.

Koay, T. B., Chitre, M. (2013) *Energy-efficient path planning for fully propelled AUVs in congested coastal waters.* Oceans MTS/IEEE Bergen: The Challenges of the Northern Dimension, 2013.

Kwok, K. S., Driessen, B. J., Phillips, C., Tovey, C. A. (2002) Analyzing the multiple-target-multiple-agent scenario using optimal assignment algorithms. Journal of Intelligent and Robotic Systems, 2002, 35(1):111-122.

MahmoudZadeh, S., Powers, D., Sammut, K., Yazdani, A. (2015) *Optimal Route Planning with Prioritized Task Scheduling for AUV Missions*. IEEE International Symposium on Robotics and Intelligent Sensors, 2015, pp.7-15.

MahmoudZadeh, S., Powers, D., Yazdani, A. (2016 *a*) *A Novel Efficient Task-Assign Route Planning Method for AUV Guidance in a Dynamic Cluttered Environment*. IEEE Congress on Evolutionary Computation (CEC), Vancouver, Canada, 2016, pp.678-684. CoRR abs/1604.02524.

MahmoudZadeh, S., Powers, D., Sammut, K., Yazdani, A. (2016 *b*) *A Novel Versatile Architecture for Autonomous Underwater Vehicle's Motion Planning and Task Assignment*. Journal of Soft Computing. 2016, 20(188):1-24. DOI 10.1007/s00500-016-2433-2

MahmoudZadeh, S., Powers, D., Sammut, K., Yazdani, A. (2016 *c*)*Biogeography-based combinatorial strategy for efficient autonomous underwater vehicle motion planning and task-time management.* Journal of Marine Science and Application, Springer, 2016, Vol. 15 (4): 463-477.

MahmoudZadeh, S., Yazdani, A., Sammut, K., Powers, D. (2017-*a*) *Online Path Planning for AUV Rendezvous in Dynamic Cluttered Undersea Environment Using Evolutionary Algorithms.* Journal of Applied Soft Computing, 2017. doi.org/10.1016/j.asoc.2017.10.025

MahmoudZadeh, S., Powers, D., Sammut, K. (2017-*b*) *An Autonomous Dynamic Motion-Planning Architecture for Efficient AUV Mission Time Management in Realistic Sever Ocean Environment*. Robotics and Autonomous Systems, 2017, vol. 87, pp. 81-103.

MahmoudZadeh, S., Powers, D., Sammut, K., Yazdani, A. (2018-*a*) Efficient AUV Path Planning in Time-Variant Underwater Environment Using Differential Evolution Algorithm. ***Journal of Marine Science and Application***, 2018. DOI: 10.1007/s11804-018-0034-4

MahmoudZadeh, S., Powers, D., Sammut, K., Atyabi, A., Yazdani, A. (2018-b) Hybrid Motion Planning Task Allocation Model for AUV's Safe Maneuvering in a Realistic Ocean Environment. ***Journal of Intelligent & Robotic Systems***, ***Springer***, 2018, pp. 1-18.

MahmoudZadeh, S., Powers, D., Sammut, K., Atyabi, A., Yazdani, A. (2018-c) A hierarchal planning framework for AUV mission management in a spatiotemporal varying ocean. ***Journal of Computers & Electrical Engineering***, ***Elsevier***, 2018, Vol. 67, pp. 741-760

MahmoudZadeh, S., Powers, K., Atyabi A., (2018-*d*) UUV's Hierarchical DE-based Motion Planning in a Semi Dynamic Underwater Wireless Sensor Network. *IEEE Transactions on Cybernetics*. 99:1-14. DOI: 10.1109/TCYB.2018.2837134

S MahmoudZadeh, DMW Powers, K Sammut, AM Yazdani, A Atyabi. (2018-*e*) Hybrid Motion Planning Task Allocation Model for AUV's Safe Maneuvering in a Realistic Ocean Environment. Journal of Intelligent & Robotic Systems, Springer, 2018, pp. 1-18.

MahmoudZadeh, S., Powers, D., Bairam Zadeh, R., (2019) Autonomy and Unmanned Vehicles "Augmented Reactive Mission–Motion Planning Architecture for Autonomous Vehicles", Springer Nature, Cognitive Science and Technology, ISBN 978-981-13-2245-7. DOI: 10.1007/978-981-13-2245-7

Petres, C., Pailhas, Y., Patron, P., Petillot, Y., Evans, J., Lane, D. (2007) *Path planning for autonomous underwater vehicles*. IEEE Transactions on Robotics, 2007, vol. 23, no.2, pp. 331-341.

Roberge V, Tarbouchi M, Labonte G. (2013) *Comparison of Parallel Genetic Algorithm and Particle Swarm Optimization for Real-Time UAV Path Planning.* IEEE Trans on Industrial Informatics, 2013, vol. 9, no.1, pp. 132-141.

Yang, X. S. (2010) *Nature-Inspired Metaheuristic Algorithms*. Luniver Press, UK. 2nd edition, 2010.

Yang, X. S., He, X. (2013) *Firefly Algorithm: Recent Advances and Applications*. Journal of Swarm Intelligence, 2013, vol.1, no.1, pp.36-50.